\documentclass[11pt,a4paper]{article}
\usepackage[hyperref]{emnlp-ijcnlp-2019}
\usepackage{times}
\usepackage{graphicx}
\usepackage{xcolor}
\usepackage{float}
\usepackage{booktabs}
\usepackage{latexsym}
\usepackage{pbox}
\usepackage{multirow}
\usepackage[utf8]{inputenc}
\usepackage[frenchb, english]{babel}
\usepackage{subfig}
\usepackage{todonotes}
\usepackage{tikz-qtree}
\usepackage{tikz-qtree-compat}

\newcommand{\key}[1]{\textsc{#1}}
\usepackage{url}

\usepackage{linguex}
\usepackage{tikz}

\aclfinalcopy 

\title{Representation of Constituents in Neural Language Models:\\ Coordination Phrase as a Case Study}

\author{
Aixiu An$^1$ \quad Peng Qian$^2$ \quad Ethan Wilcox$^3$ \quad Roger Levy$^2$ \\
  $^1$ Université de Paris, LLF, CNRS, {\tt aixiu.an@etu.univ-paris-diderot.fr} \\
  $^2$ Department of Brain and Cognitive Sciences, MIT, {\tt \{pqian, rplevy\}@mit.edu}  \\
  $^3$ Department of Linguistics, Harvard University, {\tt wilcoxeg@g.harvard.edu}
  }

\date{}

\begin{document}
\maketitle
\begin{abstract}

Neural language models have achieved state-of-the-art performances on many NLP tasks, and recently have been shown to learn a number of hierarchically-sensitive syntactic dependencies between individual words. However, equally important for language processing is the ability to combine words into phrasal constituents, and use constituent-level features to drive downstream expectations. Here we investigate neural models' ability to represent constituent-level features, using coordinated noun phrases as a case study. We assess whether different neural language models trained on English and French represent phrase-level number and gender features, and use those features to drive downstream expectations. Our results suggest that models use a linear combination of NP constituent number to drive CoordNP/verb number agreement. This behavior is highly regular and even sensitive to local syntactic context, however it differs crucially from observed human behavior. Models have less success with gender agreement. Models trained on large corpora perform best, and there is no obvious advantage for models trained using explicit syntactic supervision. 

\end{abstract}

\section{Introduction}

Humans deploy structure-sensitive expectations to guide processing during natural language comprehension \cite{levy2008expectation}. While it has been shown that neural language models show similar structure-sensitivity in their predictions about upcoming material \cite{linzen2016assessing, futrell2018}, previous work has focused on dependencies that are conditioned by features attached to a single word, such as subject number \cite{gulordava2018, marvin2018targeted} or wh-question words \cite{wilcox2018}. There has been no systematic investigation into models' ability to compute phrase-level features---features that are attached to a set of words---and whether models can deploy these more abstract properties to drive downstream expectations.

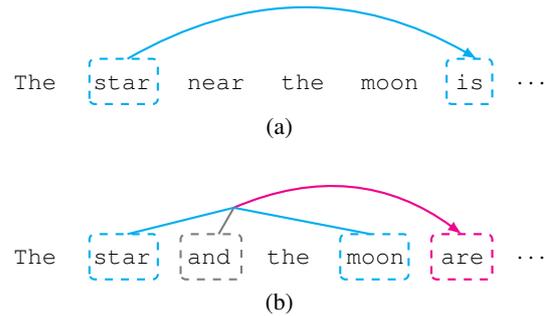
\begin{figure}
\centering
\subfloat[]{
\begin{tikzpicture}
\path (0,0) node (sent1) {\small \texttt{The \hspace{0.1cm} star \hspace{0.1cm} near \hspace{0.1cm} the \hspace{0.1cm} moon \hspace{0.1cm} is  \hspace{0.05cm} $\cdots$ }};

\draw[cyan, thick, dashed, rounded corners=2] (-2.5,-0.3) rectangle (-1.6,0.3);
\draw[cyan, thick, dashed, rounded corners=2] (2.2,-0.3) rectangle (2.8,0.3);

\draw [thick, cyan, bend left,-latex] (-2,0.3) to  (2.6, 0.3);

\end{tikzpicture}\label{agreement-patterns-1}
}\\
\subfloat[]{
\begin{tikzpicture}

\path(0,-2) node (sent2) {\small \texttt{The \hspace{0.1cm} star \hspace{0.1cm} and \hspace{0.1cm} the \hspace{0.1cm} moon \hspace{0.1cm} are \hspace{0.05cm} $\cdots$ }};

\draw[cyan, thick, dashed, rounded corners=2] (-2.5,-2.3) rectangle (-1.6,-1.7);
\draw[gray, thick, dashed, rounded corners=2] (-1.3,-2.3) rectangle (-0.5,-1.7);
\draw[cyan, thick, dashed, rounded corners=2] (0.8,-2.3) rectangle (1.7,-1.7);
\draw[magenta, thick, dashed, rounded corners=2] (2,-2.3) rectangle (2.8,-1.7);

\draw [thick, magenta, bend left,-latex] (-0.6,-1.35) to  (2.4, -1.7);

\draw [thick, cyan] (-0.6,-1.35) to  (-2, -1.7);
\draw [thick, gray] (-0.6,-1.35) to  (-0.8, -1.7);
\draw [thick, cyan] (-0.6,-1.35) to  (1.2, -1.7);
\end{tikzpicture}\label{agreement-patterns-2}
}
\caption{Subject-verb agreement with (a) the head of a noun phrase structure, and (b) the coordination structure.}\label{agreement-patterns}
\end{figure}

In this work, we assess whether state-of-the-art neural models can compute and employ phrase-level gender and number features of coordinated subject Noun Phrases (CoordNPs) with two nouns. Typical syntactic phrases are \key{endocentric}: they are \key{headed} by a single child, whose features determine the agreement requirements for the entire phrase. In Figure~\ref{agreement-patterns-1}, for example, the word \emph{star} heads the subject NP \emph{The star}; since \emph{star} is singular, the verb must be singular. CoordNPs lack endocentricity: neither conjunct NP solely determines the features of the NP as a whole.  Instead, these feature values are determined by compositional rules sensitive to the features of the conjuncts and the identity of the coordinator. In Figure \ref{agreement-patterns-2}, because the coordinator is \emph{and}, the subject NP number is plural even though both conjuncts (\emph{the star} and \emph{the moon}) are singular. As this case demonstrates, the agreement behavior for CoordNPs must be driven by more abstract, constituent-level representations, and cannot be reduced to features hosted on a single lexical item.

We use four suites of experiments to assess whether neural models are able to build up phrase-level representations of CoordNPs on the fly and deploy them to drive humanlike behavior. First, we present a simple control experiment to show that models can represent number and gender features of non-coordinate NPs \textbf{(Non-coordination Agreement)}. Second, we show that models modulate their expectations for downstream verb number based on the CoordNP's coordinating conjunction combined with the features of the coordinated nouns \textbf{(Simple Coordination)}. We rule out the possibility that models are using simple heuristics by designing a set of stimuli where a simple heuristic would fail due to structural ambiguity \textbf{(Complex Coordination)}. The striking success for all models in this experiment indicates that even neural models with no explicit hierarchical bias, trained on a relatively small amount of text are able to learn fine-grained and robust generalizations about the interaction between CoordNPs and local syntactic context. Finally, we use subject--auxiliary inversion to test whether an upstream lexical item modulates model expectation for the phrasal-level features of a downstream CoordNP \textbf{(Inverted Coordination)}. Here, we find that all models are insensitive to the fine-grained features of this particular syntactic context. Overall, our results indicate that neural models can learn fine-grained information about the interaction of Coordinated NPs and local syntactic context, but their behavior remains unhumanlike in many key respects.

\section{Methods}

\subsection{Psycholinguistics Paradigm}

To determine whether state-of-the-art neural architectures are capable of learning humanlike CoordNP/verb agreement properties, we adopt the psycholinguistics paradigm for model assessment. In this paradigm the models are tested using hand-crafted sentences designed to test underlying network knowledge. The assumption here is that if a model implicitly learns humanlike linguistic knowledge during training, its expectations for upcoming words should qualitatively match human expectations in novel contexts. For example, \citet{linzen2016assessing} and \citet{kuncoro2016recurrent} assessed how well neural models had learned the subject/verb number agreement by feeding them with the prefix \textit{The keys to the cabinet ...}. If the models predicted the grammatical continuation \textit{are} over the ungrammatical continuation \textit{is}, they can be said to have learned the number agreement insofar as the number of the head noun and not the number of the distractor noun, \textit{cabinet}, drives expectations about the number of the matrix verb. 

If models are able to robustly modulate their expectations based on the internal components of the CoordNP, this will provide evidence that the networks are building up a context-sensitive phrase-level representation. We quantify model expectations as \key{surprisal values}. Surprisal is the negative log-conditional probability $S(x_i) = -\log_2 p(x_i|x_1 \dots x_{i-1})$ of a sentence's $i^{th}$ word $x_i$ given the previous words. Surprisal tells us how strongly $x_i$  is expected in context and is known to correlate with human processing difficulty \citep{hale2001probabilistic,levy2008expectation,smith2013effect}. In the CoordNP/Verb  agreement studies presented here, cases where the proceeding context sets high expectation for a number-inflected verb form $w_i$, (e.g. singular `is') we would expect $S(w_i)$ to be lower than its number-mismatched counterpart (e.g. plural `are').

\begin{table}[t]
\small
    \centering
    \begin{tabular}{|c|l|l|r|}
    \hline
 & \multicolumn{1}{|c|}{Model} &  \multicolumn{1}{|c|}{Training data} & \multicolumn{1}{|c|}{\# tokens}\\\hline
    \multirow{5}{*}{\rotatebox[origin=c]{90}{English}} & LSTM (PTB) & Penn Treebank & $\sim$ 1M\\ 
        & ActionLSTM (PTB) & Penn Treebank & $\sim$ 1M\\ 
        & RNNG (PTB) & Penn Treebank  & $\sim$ 1M \\ 
        & LSTM (enWiki) & English Wikipedia & $\sim$ 90M\\ 
        & LSTM (1B) & 1 Billion Word & $\sim$ 800M\\\hline
\multirow{4}{*}{\rotatebox[origin=c]{90}{French}} & LSTM (FTB) & French Teebank & $\sim$ 0.5M \\
        & ActionLSTM (FTB) & French Teebank & $\sim$ 0.5M  \\
        & RNNG (FTB) & French Teebank & $\sim$ 0.5M  \\
        & LSTM (frWaC) & Subset of frWaC  & $\sim$ 138M \\
    \hline
    \end{tabular}
    \caption{A summary of models tested.}
    \label{tab:my_label}
\end{table}

\subsection{Models Tested}

\paragraph{Recurrent Neural Network (RNN) Language Models} are trained to output the probability distribution of the upcoming word given a context, without explicitly representing the structure of the context \cite{sundermeyer2012lstm, elman1990finding}. We trained two two-layer recurrent neural language models with long short-term memory architecture \cite{hochreiter1997long} on a relatively small corpus. The first model, referred as `{\bf LSTM (PTB)}' in the following sections, was trained on the sentences from Penn Treebank \cite{marcus19building}. The second model, referred as `{\bf LSTM (FTB)}', was trained on the sentences from French Treebank \cite{abeille2003building}. We set the size of input word embedding and LSTM hidden layer of both models as 256. 

We also compare LSTM language models trained on large corpora. We incorporate two pretrained English language models: one trained on the Billion Word benchmark (referred as `\textbf{LSTM (1B)}') from \citet{jozefowicz2016}, and the other trained on English Wikipedia (referred as `\textbf{LSTM (enWiki)}') from \citet{gulordava2018}. For French, we trained a large LSTM language model (referred as `\textbf{LSTM (frWaC)}') on a random subset (about 4 million sentences, 138 million word tokens) of the frWaC dataset \cite{baroni2009wacky}. 
We set the size of the input embeddings and hidden layers to 400 for the LSTM (frWaC) model since it is trained on a large dataset.

\paragraph{ActionLSTM} models the linearized bracketed tree structure of a sentence by learning to predict the next action required to construct a phrase-structure parse \cite{choe-charniak-2016-parsing}. The action space consists of three possibilities: open a new non-terminal node and opening bracket; generate a terminal node; and close a bracket. To compute surprisal values for a given token, we approximate $P(w_i|w_{1\cdots i-1)}$ by marginalizing over the most-likely partial parses found by word-synchronous beam search \cite{stern2017effective}.

\paragraph{Generative Recurrent Neural Network Grammars (RNNG)} jointly model the word sequence as well as the underlying syntactic structure \cite{dyer2016recurrent}. Following  \citet{hale2018finding}, we estimate surprisal using word-synchronous beam search \cite{stern2017effective}. We use the same hyper-parameter settings as \citet{dyer2016recurrent}.

The annotation schemes used to train the syntactically-supervised models differ slightly between French and English. In the PTB (English) CoordNPs are flat structures bearing an `NP' label. In FTB (French), CoordNPs are binary-branching, labeled as NPs, except for the phrasal node dominating the coordinating conjunction, which is labeled `COORD'. We examine the effects of annotation schemes on model performance in Appendix \ref{appx:annotation}. \footnote{The materials and code for this project can be found in \url{https://github.com/cpllab/rnn\_psycholing\_coordination.git}}

\section{Experiment 1: Non-coordination Agreement} \label{sec:control}

In order to provide a baseline for following experiments, here we assess whether the models tested have learned basic representations of number and gender features for non-coordinated Noun Phrases. We test number agreement in English and French as well as gender agreement in French. Both English and French have two grammatical number feature: {\sc singular} (sg) and {\sc plural} (pl). French has two grammatical gender features: {\sc masculine} (m) and {\sc feminine} (f).

The experimental materials include sentences where the subject NPs contain a single noun which can either match with the matrix verb (in the case of number agreement) or a following predicative adjective (in the case of gender agreement). Conditions are given in Table \ref{tab:conditionexp1} and Table \ref{tab:conditionexp1gender}. We measure model behavior by computing the \textit{plural expectation}, or the surprisal of the singular continuation minus the surprisal of the plural continuation for each  condition and took the average for each condition. We expect a positive \textit{plural expectation} in the \textit{Npl} conditions and a negative \textit{plural expectation} in the \textit{Nsg} conditions. For gender expectation we compute a \textit{gender expectation}, which is S(feminine continuation) $-$ S(masculine continuation). We measure surprisal at the verbs and predicative adjectives themselves. 

\begin{table}[t]
\centering
\small
\begin{tabular}{cl}
\toprule
{} Condition & Sentence   \\
\midrule
Npl & The windows is/are\\
Nsg & The window is/are\\
\bottomrule
 \end{tabular}
 \caption{Conditions of number agreement in Non-coordination Agreement experiment.}
 \label{tab:conditionexp1}
 \end{table}
 
 \begin{table}[t]
\centering
\small
\begin{tabular}{cl}
\toprule
{} Condition & Sentence   \\
\midrule
\multirow{2}{*}{Nm} & Les coûts sont importants/importantes\\
& the cost.{\sc mpl} are important.{\sc mpl/fpl} \\
\multirow{2}{*}{Nf} & Les dépenses sont importants/importantes\\
&  the 
expense.{\sc fpl} are important.{\sc mpl/fpl} \\
\bottomrule
 \end{tabular}
 \caption{Conditions of gender agreement in Non-coordination Agreement experiment.}
 \label{tab:conditionexp1gender}
 \end{table}
 
\begin{figure}[!t]
    \centering
    \subfloat[English number agreement]{
    \includegraphics[ width=\linewidth]{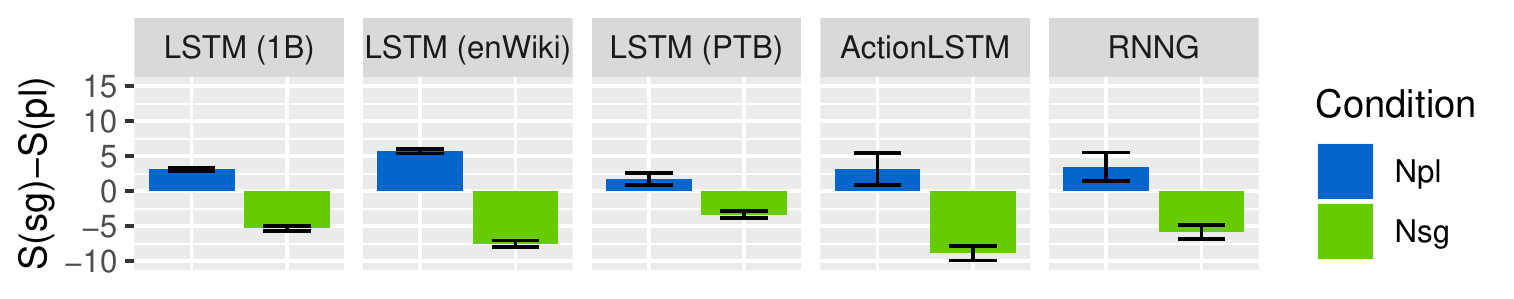}\label{control_num-en}
    }
    
    \subfloat[French number agreement]{
    \includegraphics[ width=0.96\linewidth]{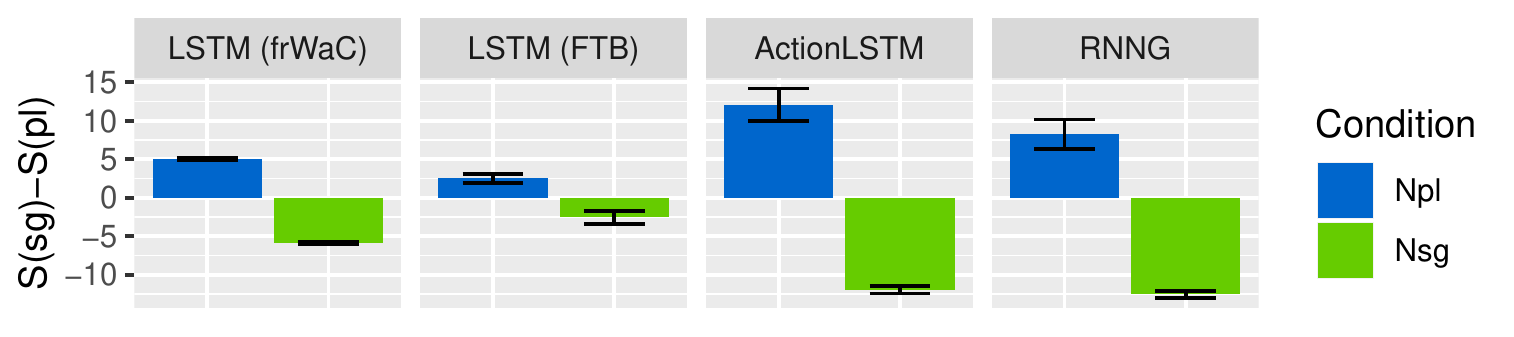}\label{control_num-fr}
    }
    
    \subfloat[French gender agreement]{
    \includegraphics[ width=0.96\linewidth]{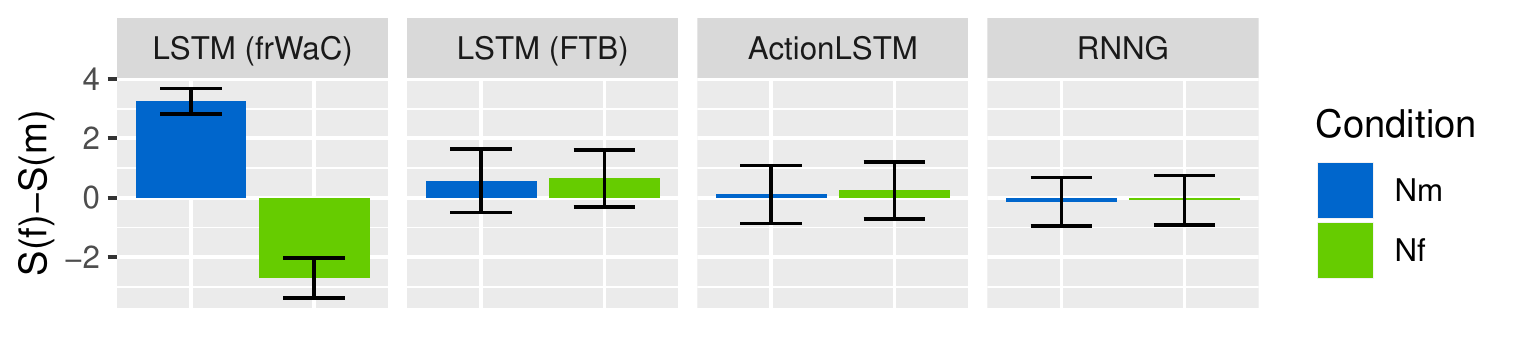}\label{control_gender-fr}
    }
    \caption{Non-Coordination Agreement experiments for English (number) and French (number and gender).}
    \label{control-coord}
\end{figure}

The results for this experiment are in Figure \ref{control-coord}, with the \textit{plural expectation} and \textit{gender expectation} on the y-axis and conditions on the x-axis. For this and subsequent experiments error bars represent 95\% confidence intervals for across-item means. For number agreement, all the models in English and French show positive plural expectation when the head noun is plural and negative plural expectation when it is singular. For gender agreement, however, only the LSTM (frWaC) shows modulation of gender expectation based on the gender of the head noun. This is most likely due to the lower frequency of predicative adjectives compared to matrix verbs in the corpus.

\section{Experiment 2: Simple Coordination}

\begin{figure*}
    \centering
    \subfloat[English and-coordination]{
    \includegraphics[ width=0.48\linewidth]{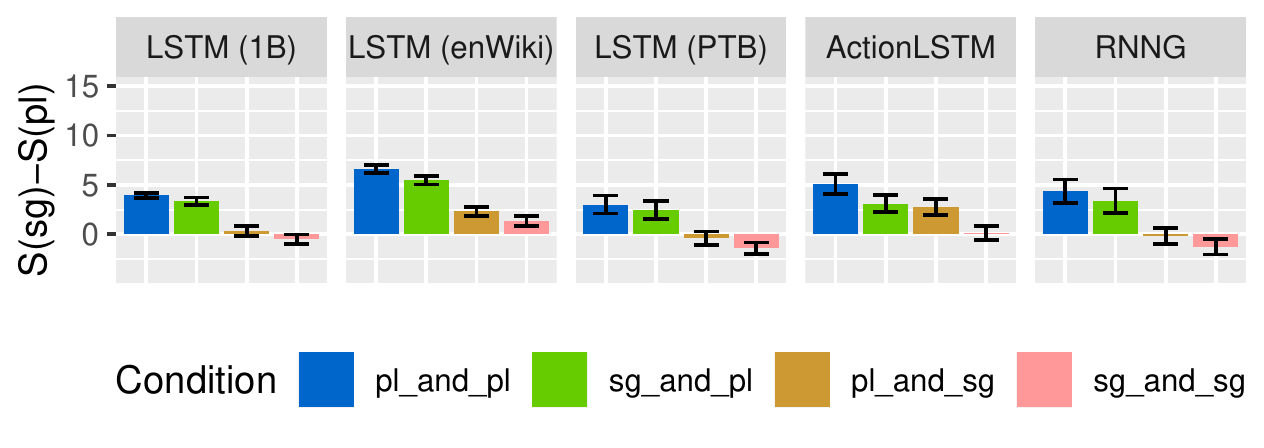}\label{simple-coord-and-en}
    }
    \subfloat[French and-coordination]{
    \includegraphics[ width=0.48\linewidth]{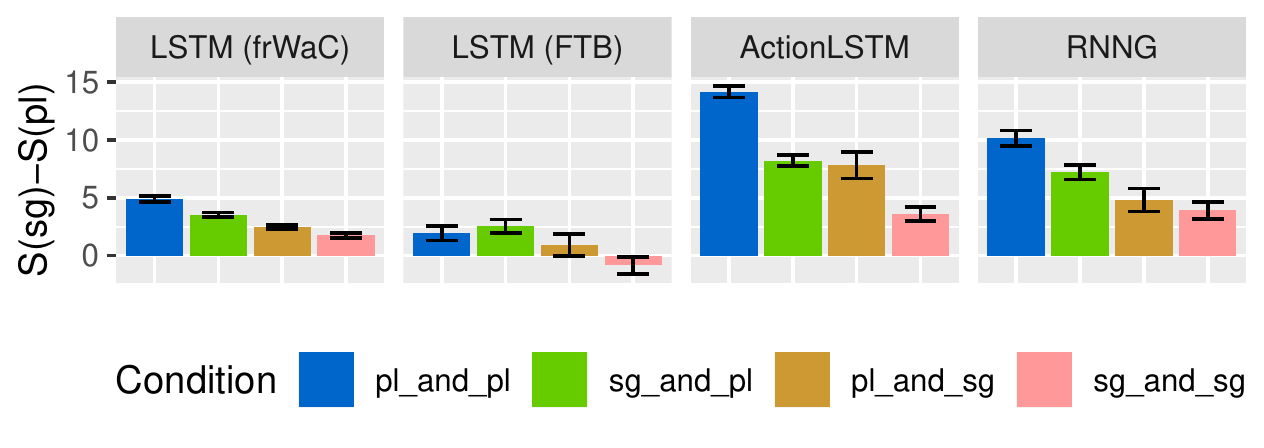}\label{simple-coord-and-fr}
    }\\
    \subfloat[English or-coordination]{
    \includegraphics[ width=0.48\linewidth]{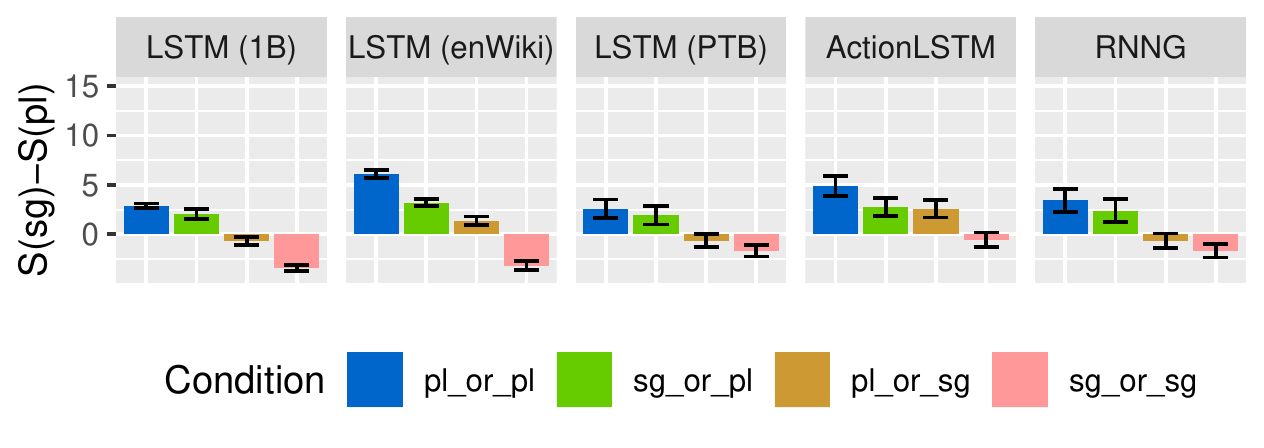}\label{simple-coord-or-en}
    }
    \subfloat[French or-coordination]{
    \includegraphics[ width=0.48\linewidth]{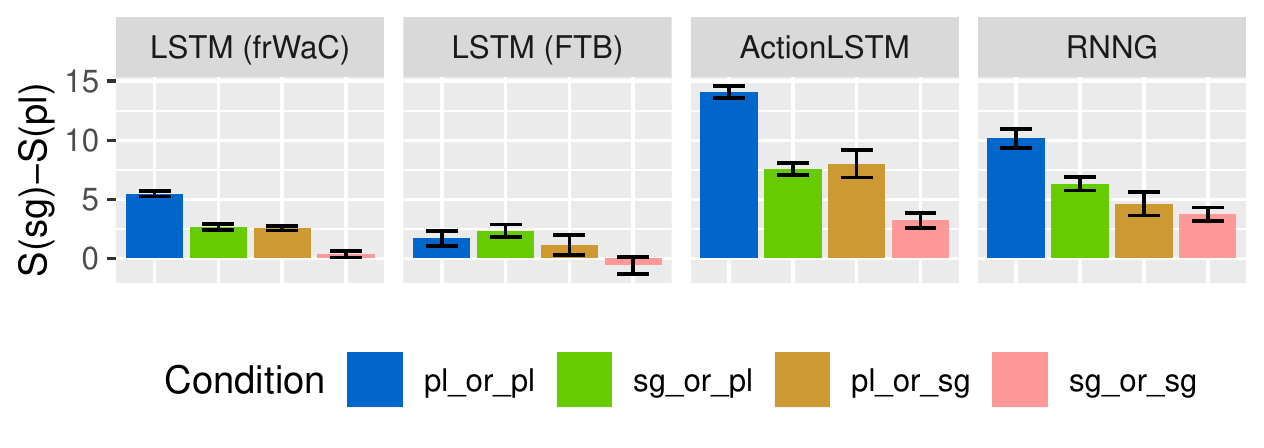}\label{simple-coord-or-fr}
    }
    \caption{Comparison of models' expectation preferences for singular vs. plural predicate in English and French Simple Coordination experiments.}
    \label{simple-coord}
\end{figure*}

In this section, we test whether neural language models can use grammatical features hosted on multiple components of a coordination phrase---the coordinated nouns as well as the coordinating conjunction---to drive downstream expectations. We test number agreement in both English and French and gender agreement in French.

\subsection{Number Agreement}\label{sec:simple-number}

In simple subject/verb number agreement, the number features of the CoordNP are determined by the coordinating conjunction and the number features of the two coordinated NPs. CoordNPs formed by \emph{and} are plural and thus require plural verbs; CoordNPs formed by \emph{or}  allow either plural or singular verbs, often with the number features of the noun linearly closest to the verb playing a more important role, although this varies cross-linguistically \cite{fowler1992}. Forced-choice preference experiments in \citet{keung2018} reveal that English native speakers prefer singular agreement when the closest conjunct in an \emph{or}-CoordNP is singular and plural agreement when the closest conjunct is plural. In French, both singular and plural verbs are possible when two singular NPs are joined via disjunction \citep{goosse2016}. 

In order to assess whether the neural models learn the basic CoordNP licensing for English, we adapted 37 items from \citet{keung2018}, along the 16 conditions outlined in Table \ref{tab:conditionexp-simple-number}. Test items consist of the sentence preamble, followed by either the singular or plural \textit{BE} verb, half the time in present tense (\textit{is/are}) and half the time in past tense \textit{(was/were)}. We measured the plural expectation, following the procedure in Section \ref{sec:control}. We created 24 items using the same conditions as the English experiment to test the models trained in French, using the 3$^{rd}$ person  singular and plural form of verb \textit{aller}, `to go' (\textit{va, vont}). Within each item, nouns match in gender; across all conditions half the nouns are masculine, half feminine.

\begin{table}[t]
\centering
\small
\begin{tabular}{ll}
\toprule
{} Condition & Sentence  \\
\midrule
pl\_and\_pl & The doors and the windows is/are\\
sg\_and\_pl&  The door and the windows is/are\\
pl\_and\_sg&  The doors and the window is/are\\
sg\_and\_sg & The door and the window is/are\\
pl\_or\_pl & The doors or the windows is/are\\
sg\_or\_pl&  The door or the windows is/are\\
pl\_or\_sg& The doors or the window is/are\\
sg\_or\_sg & The door or the window is/are\\
\bottomrule
 \end{tabular}
 \caption{Conditions of number agreement in Simple Coordination experiment.}
 \label{tab:conditionexp-simple-number}
 \end{table}

The results for this experiment can be seen in Figure \ref{simple-coord}, with the results for English on the left and French on the right. The results for \textit{and} are on the top row, \textit{or} on the bottom row. For all figures the y-axis shows the plural expectation, or the difference in surprisal between the \textit{singular} condition and the \textit{plural} condition. Turning first to \textbf{English-\textit{and}} (Figure \ref{simple-coord-and-en}), all models show plural expectation (the bars are significantly greater than zero) in the \textit{pl\_and\_pl} and \textit{sg\_and\_pl} conditions, as expected. For the \textit{pl\_and\_sg} condition, only the LSTM (enWiki) and ActionLSTM are greater than zero, indicating humanlike behavior. For the \textit{sg\_and\_sg} condition, only the LSTM (enWiki) model shows the correct plural expectation. For the \textbf{French-\textit{and}} (Figure \ref{simple-coord-and-fr}), all models show positive plural expectation in all conditions, as expected, except for the LSTM (FTB) in the \textit{sg\_and\_sg} condition.

Examining the results for \textbf{English-\textit{or}}, we find that all models demonstrate humanlike expectation in the \textit{pl\_or\_pl} and \textit{sg\_or\_pl} conditions. The LSTM (1B), LSTM (PTB), and RNNG models show zero or negative singular expectation for the \textit{pl\_or\_sg} conditions, as expected. However the LSTM (enWiki) and ActionLSTM models show positive plural expectation in this condition, indicating that they have not learned the humanlike generalizations. All models show significantly negative plural expectation in the \textit{sg\_or\_sg} condition, as expected. In the \textbf{French-\textit{or}} cases, models show almost identical behavior to the \textit{and} conditions, except the LSTM (frWaC) shows smaller plural expectation when singular nouns are linearly proximal to the verb.  

These results indicate moderate success at learning coordinate NP agreement, however this success may be the result of an overly simple heuristic. It appears that expectation for both plural and masculine continuations are driven by a linear combination of the two nominal number/gender features transferred into log-probability space, with the earlier noun mattering less than the later noun. A model that optimally captures human grammatical preferences should show no or only slight difference across conditions in the surprisal differential for the \emph{and} conditions, and be greater than zero in all cases. Yet, all the models tested show gradient performance based on the number of plural conjuncts.

\subsection{Gender Agreement} \label{sec:gender-agreement}

In French, if two nouns are coordinated with \textit{et} (\textit{and}-coordination), agreement must be masculine if there is one masculine element in the coordinate structure. If the nouns are coordinated with \textit{ou} (\textit{or}-coordination), both masculine and feminine agreement is acceptable \cite{corbett1991, wechsler2003}. Although linear proximity effects have been tested for a number of languages that employ grammatical gender, as in e.g. Slavic languages \citep{willer2018}, there is no systematic study for French. 

\begin{table}[!th]
\centering
\scriptsize
\begin{tabular}{ll}
\toprule
\multicolumn{1}{c}{Condition} & \multicolumn{1}{c}{Sentence}  \\
\midrule
\multirow{2}{*}{m\_and\_m} & \pbox{6cm}{Les prix et les coûts sont importants/importantes}\\
& the price.{\sc mpl} and the cost.{\sc mpl} are important.{\sc mpl/fpl} \\
\multirow{2}{*}{f\_and\_m} &  \pbox{6cm}{Les recettes et les coûts sont importants/importantes} \\
& the
revenues.{\sc fpl} and the cost.{\sc mpl} are important.{\sc mpl/fpl} \\
\multirow{2}{*}{m\_and\_f} & \pbox{6cm}{Les prix et les dépenses sont importants/importantes}\\
& the price.{\sc mpl} and the expense.{\sc fpl} are important.{\sc mpl/fpl}\\
\multirow{2}{*}{f\_and\_f} &  \pbox{6cm}{Les recettes et les dépenses sont importants/importantes} \\
& the revenues.{\sc fpl} and the expense.{\sc fpl} are important.{\sc mpl/fpl}  \\
\bottomrule
 \end{tabular}
 \caption{Conditions for the and-coordination experiment. 
 (Items for or-coordination are the same except that we change the coordinator to  \textit{ou}.)
 }
 \label{tab:gender1}
 \end{table}

To assess whether the French neural models learned humanlike gender agreement, we created 24 test items, following the examples in Table \ref{tab:gender1}, and measured the masculine expectation. In our test items, the coordinated subject NP is followed by a predicative adjective, which either takes on masculine or feminine gender morphology.
 
\begin{figure}
    \centering
    \subfloat[French and-coordination]{
    \includegraphics[ width=\linewidth]{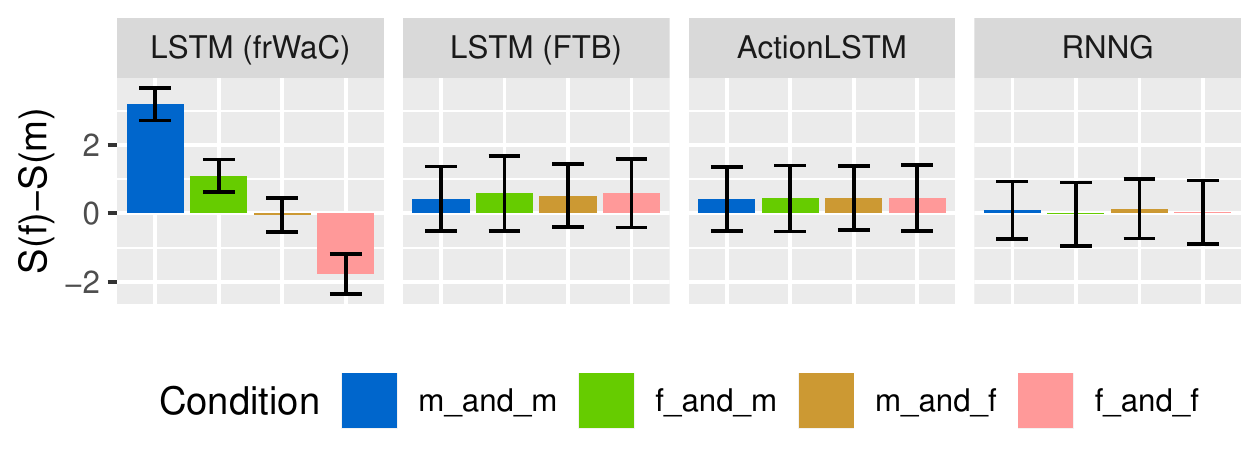}
    }
    
    \subfloat[French or-coordination]{
    \includegraphics[ width=\linewidth]{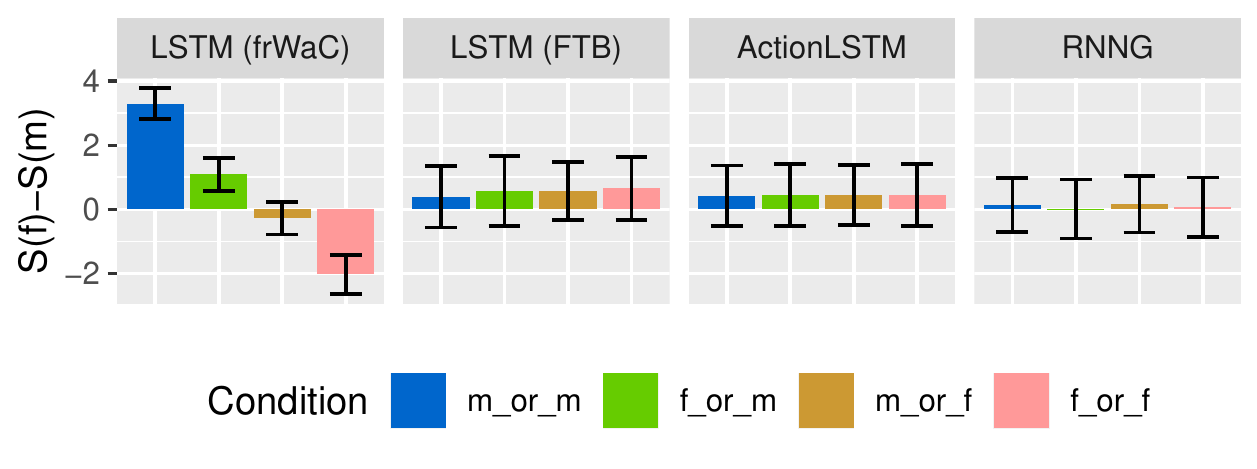}
    }
    \caption{Comparison of models' expectation preferences for Feminine v.s. Masculine predicative adjectives in French.}
    \label{simple-coord-gender}
\end{figure}

Results from the experiment can be seen in Figure \ref{simple-coord-gender}. No models shows qualitative difference based on the coordinator, and only the LSTM (frWaC) shows significant behavior difference between conditions. Here, we find positive masculine expectation in the \textit{m\_and\_m} and \textit{f\_and\_m} conditions, and negative masculine expectation in the \textit{f\_and\_f} condition, as expected. However, in the \textit{m\_and\_f} condition, the masculine expectation is not significantly different from zero, where we would expect it to be positive. In the \textit{or}-coordination conditions, following our expectation, masculine expectation is positive when both conjuncts are masculine and negative when both are feminine. For the LSTM (FTB) and ActionLSTM models, the masculine expectation is positive (although not significantly so) in all conditions, consistent with results in Section \ref{sec:control}.

\section{Experiment 3: Complex Coordination}

One possible explanation for the results presented in the previous section is that the models are using a `bag of features' approach to plural and masculine licensing that is opaque to syntactic context: Following a coordinating conjunction surrounded by nouns, models simply expect the following verb to be plural, proportionally to the number of plural nouns. 

In this section, we control for this potential confound by conducting two experiments: In the \textit{Complex Coordination Control} experiments we assess models' ability to extend basic CoordNP licensing into sententially-embedded environments, where the CoordNP can serve as an embedded subject. In the \textit{Complex Coordination Critical} experiments, we leverage the sentential embedding environment to demonstrate that when the CoordNPs cannot plausibly serve as the subject of the embedded phrase, models are able to suppress the previously-demonstrated expectations set up by these phrases. These results demonstrate that models are not following a simple strategy for predicting downstream number and gender features, but are building up CoordNP representations on the fly, conditioned on the local syntactic context.

\subsection{Complex Coordination Control}

Following certain sentential-embedding verbs, CoordNPs serve unambiguously as the subject of the verb's sentence complement and should trigger number agreement behavior in the main verb of the embedded clause, similar to the behavior presented in \ref{sec:simple-number}. To assess this, we use the 37 test items in English and 24 items in French in section \ref{sec:simple-number}, following the conditions in Table \ref{tab:chunk1} (for number agreement), testing only \textit{and} coordination. For gender agreement, we use the same test items and conditions for \textit{and} coordination in Section \ref{sec:gender-agreement}, but with the Coordinated NPs embedded in a context similar to \ref{gender_complex_that}. As before, we derived the plural expectation by measuring the difference in surprisal between the singular and plural continuations and the gender expectation by computing the difference in surprisal between the masculine and feminine predicates.

 \exg. Je croyais que les prix et les dépenses étaient importants/importantes.\\
 I thought that the.{\sc pl} price.{\sc mpl} and the.{\sc pl} expense.{\sc fpl} were important.{\sc mpl/fpl}\\
I thought that the prices and the expenses were important.  \label{gender_complex_that}

\begin{table}[th]
\centering
\small
\begin{tabular}{ll}
\toprule
\multicolumn{1}{c}{Condition} & \multicolumn{1}{c}{Sentence }  \\
\midrule
pl\_and\_pl & I think that the doors and the windows is/are\\
sg\_and\_pl&  I think that the door and the windows is/are\\
pl\_and\_sg&  I think that the doors and the window is/are\\
sg\_and\_sg & I think that the door and the window is/are\\
\bottomrule
 \end{tabular}
 \caption{Conditions of number agreement in Complex Coordination Control experiment.}
 \label{tab:chunk1}
 \end{table}

\begin{figure}[t]
    \centering
    \subfloat[English number agreement]{
    \includegraphics[width=\linewidth]{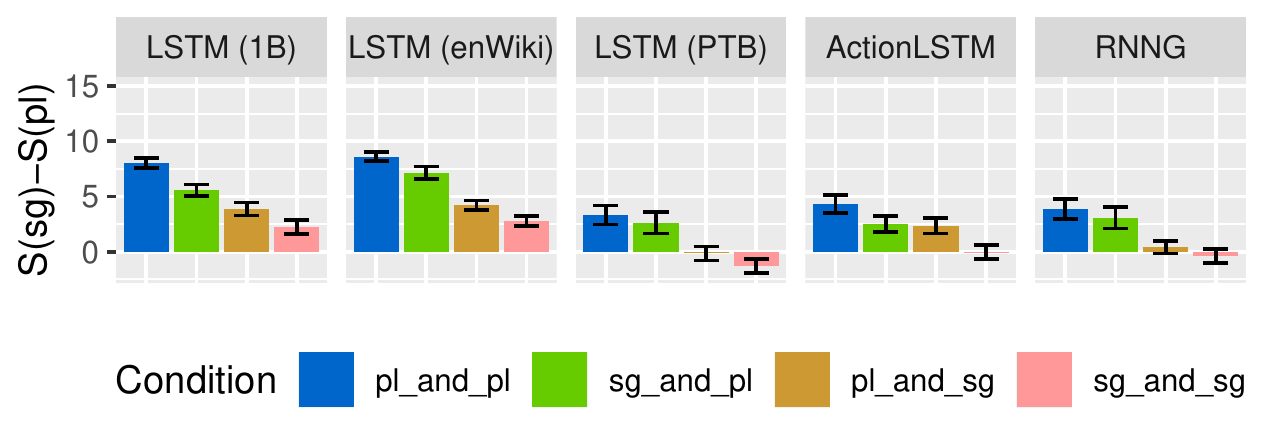}\label{that-coord-number-en}
    }\\
    \subfloat[French number agreement]{
    \includegraphics[width=\linewidth]{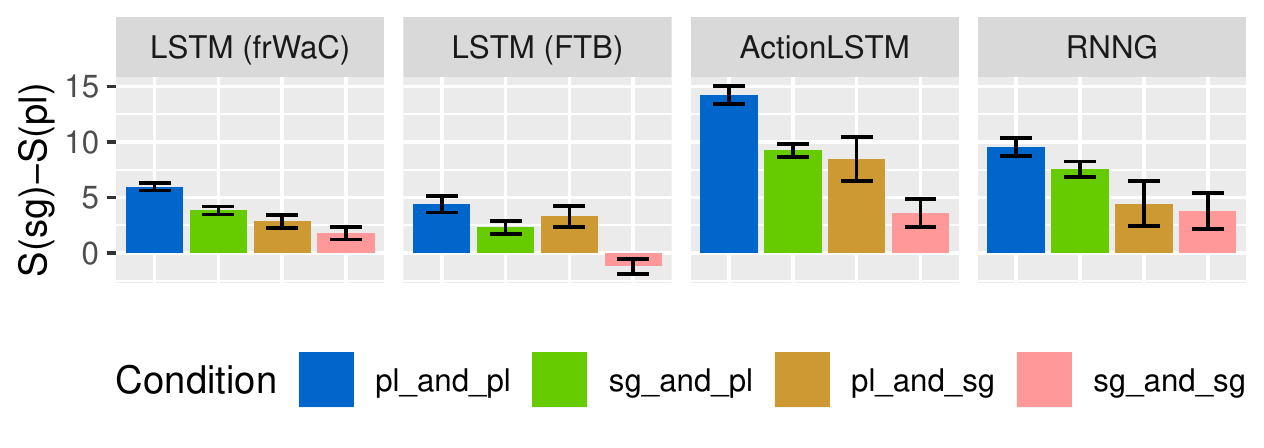}\label{that-coord-number-fr}
    }\\
    \subfloat[French gender agreement]{
    \includegraphics[width=\linewidth]{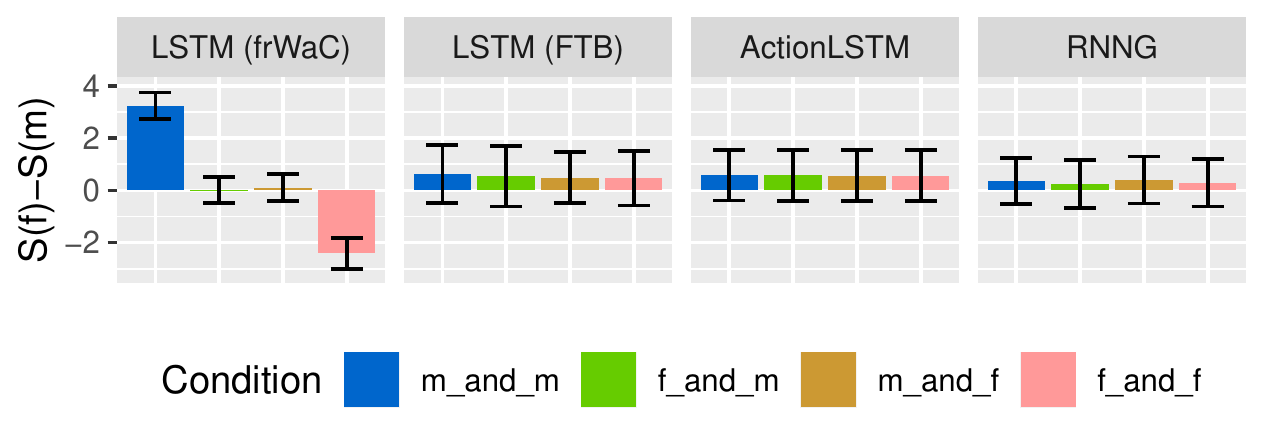}\label{that-coord-gender-fr}
    }
\caption{Comparison of model's expectation preferences in the Complex Coordination Control experiments.}\label{complex-coord-controlled}
\end{figure}

The results for the control experiments can be seen in Figure \ref{complex-coord-controlled}, with English number agreement on the top row, French number agreement in the middle row and French gender agreement on the bottom. The y-axis shows either plural or masculine expectation, with the various conditions along the x-axis. For English number agreement, we find that the models behave similarly as they do for simple coordination contexts. All models show significant plural expectation when the closest noun is plural, with only two models demonstrating plural expectation in the \textit{sg\_and\_sg} case. The French number agreement tests show similar results, with all models except LSTM (FTB) demonstrating significant plural prediction in all cases. Turning to French gender agreement, only the LSTM (frWaC) shows sensitivity to the various conditions, with positive masculine expectation in the \textit{m\_and\_m} condition and negative expectation in the \textit{f\_and\_f} condition, as expected. These results indicate that the behavior shown in Section \ref{sec:simple-number} extends to more complex syntactic environments---in this case to sentential embeddings. Interestingly, for some models, such as the LSTM (1B), behavior is \textit{more humanlike} when the CoordNP serves as the subject of an embedded sentence. This may be because the model, which has a large number of hidden states and may be extra sensitive to fine-grained syntactic information carried on lexical items \cite{futrell2018}, is using the complementizer, \textit{that}, to drive more robust expectations.
 
 \begin{figure}[t]
    \centering
    \subfloat[English number agreement]{
    \includegraphics[width=\linewidth]{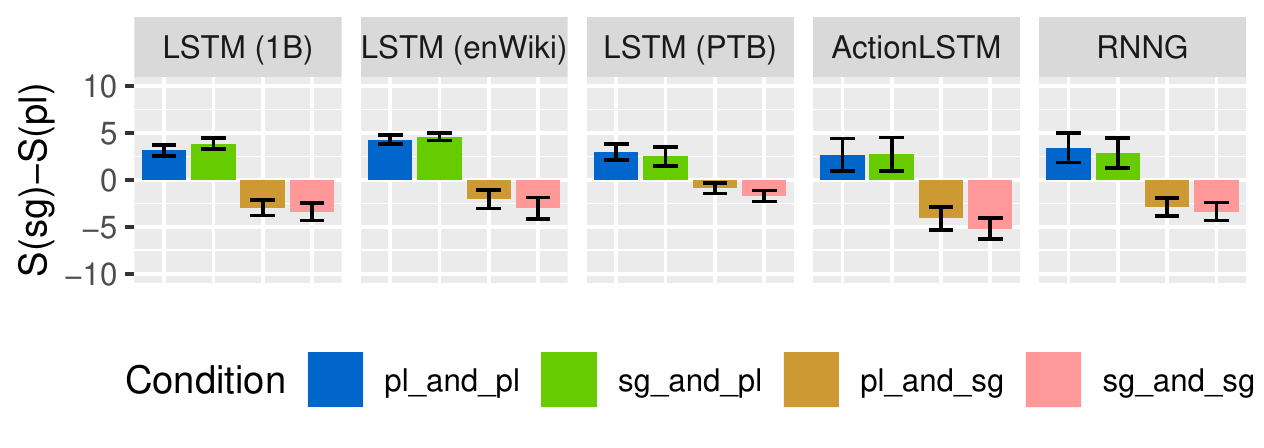}\label{verb-coord-number-en}
    }\\
    \subfloat[French number agreement]{
    \includegraphics[width=\linewidth]{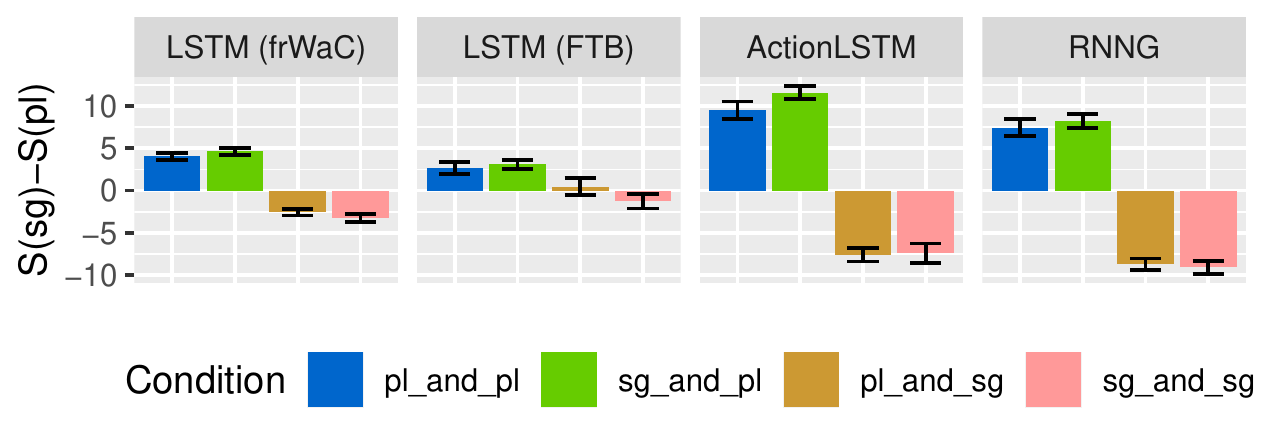}\label{verb-coord-number-fr}
    }\\
    \subfloat[French gender agreement]{
    \includegraphics[width=\linewidth]{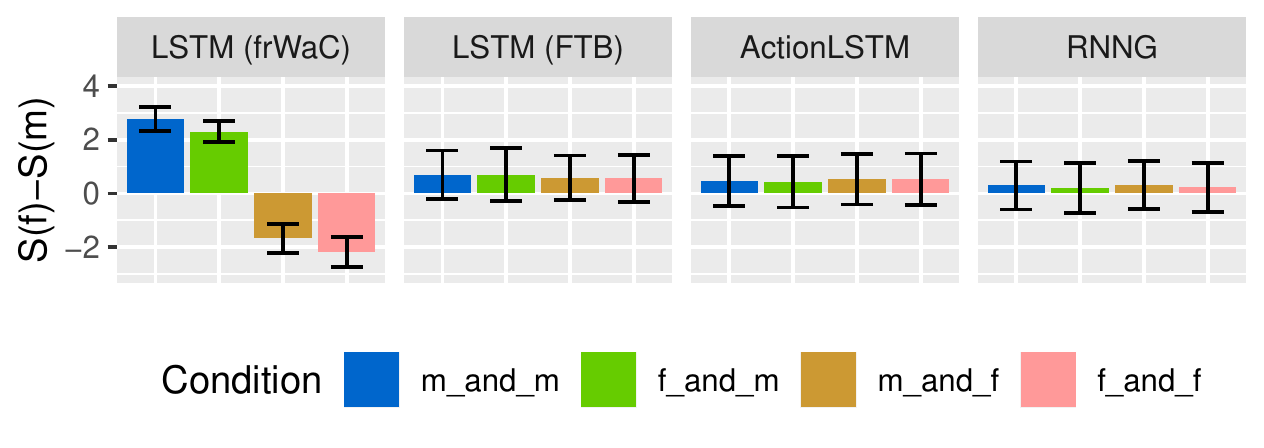}\label{verb-coord-gender-fr}
    }
\caption{Comparison of model's expectation preferences in the Complex Coordination Critical experiments.}\label{complex-coord-critical}
\end{figure}

\subsection{Complex Coordination Critical}

In order to assess whether the models' strategy for CoordNP/verb number agreement is sensitive to syntactic context, we contrast the results presented above to those from a second, critical experiment. Here, two coordinated nouns follow a verb that cannot take a sentential complement, as in the examples given in Table \ref{cond2}. Of the two possible continuations---\textit{are} or \textit{is}---the plural is only grammatically licensed when the second of the two conjuncts is plural. In these cases, the plural continuation may lead to a final sentence where the first noun serves as the verb's object and the second introduces a second main clause coordinated with the first, as in \textit{I fixed the doors and the windows are still broken.} For the same reason, the singular-verb continuation is only licensed when the noun immediately following \emph{and} is singular.

We created 37 test items in both English and French, and calculated the plural expectation. If the models were following a simple strategy to drive CoordNP/verb number agreement, then we should see either no difference in plural expectation across the four conditions or behavior no different from the control experiment.  If, however, the models are sensitive to the licensing context, we should see a contrast based solely on the number features of the second conjunct, where plural expectation is positive when the second conjunct is plural, and negative otherwise.

\begin{table}[t]
\centering
\small
\begin{tabular}{ll}
\toprule
\multicolumn{1}{c}{Condition} & \multicolumn{1}{c}{Sentence }  \\
\midrule
pl\_and\_pl & I fixed the doors and the windows is/are\\
sg\_and\_pl&  I fixed the door and the windows is/are\\
pl\_and\_sg&  I fixed the doors and the window is/are\\
sg\_and\_sg & I fixed the door and the window is/are\\
\bottomrule
 \end{tabular}
 \caption{Conditions of number agreement in Complex Coordination Critical experiment.}
 \label{cond2}
 \end{table}
 
Experimental items for a critical gender test were created similarly, as in Example \ref{gender_complex_verb}. As with plural agreement, gender expectation should be driven solely by the second conjunct: For the \textit{f\_and\_m} and \textit{m\_and\_m} conditions, the only grammatical continuation is one where the adjectival predicate bears masculine gender morphology. Conversely, for the \textit{m\_and\_f} or \textit{f\_and\_f} conditions, the only grammatical continuation is one where the adjectival predicate bears feminine morphology. As in \ref{sec:simple-number}, we created 24 test items and measured the gender expectation by calculating the difference in surprisal between the masculine and feminine continuations.

\exg. Nous avons accepté les prix et les dépenses étaient importants/importantes.\\
 we have accepted the.{\sc pl} price.{\sc mpl} and the expense.{\sc fpl} were important.{\sc mpl/fpl}\\
 We have accepted the prices and the expenses were important.  \label{gender_complex_verb}

The results from the critical experiments are in Figure \ref{complex-coord-critical}, with the English number agreement on the top row, French number agreement in the middle and gender expectation on the bottom row. Here the y-axis shows either plural expectation or masculine expectation, with the various conditions are on the x-axis. The results here are strikingly different from those in the control experiments. For number agreement, all models in both languages show strong plural expectation in conditions where the second noun is plural (blue and green bars), as they do in the control experiments. Crucially, when the second noun is singular, the plural expectation is significantly negative for all models (save for the French LSTM (FTB) \textit{pl\_and\_sg} condition). Turning to gender agreement, only the LSTM (frWaC) model shows differentiation between the four conditions tested. However, whereas the \textit{f\_and\_m} and \textit{m\_and\_f} gender expectations are not significantly different from zero in the control condition, in the critical condition they pattern with the purely masculine and purely feminine conditions, indicating that, in this syntactic context, the model has successfully learned to base gender expectation solely off of the second noun.

These results are inconsistent with a simple `bag of features' strategy that is insensitive to local syntactic context. They indicate that both models can interpret the same string as either a coordinated noun phrase, or as an NP object and the start of a coordinated VP with the second NP as its subject.

\section{Experiment 4: Inverted Coordination} \label{sec:inversion}

In addition to using phrase-level features to drive expectation about downstream lexical items, human processors can do the inverse---use lexical features to drive expectations about upcoming syntactic chunks. In this experiment, we assess whether neural models use number features hosted on a verb to modulate their expectations for upcoming CoordNPs.

To assess whether neural language models learn inverted coordination rules, we adapted items from Section \ref{sec:simple-number} in both English (37 items) and French (24 items), following the paradigm in Table \ref{cond3}. The first part of the phrase contains either a plural or singular verb and a plural or singular noun. In this case, we sample the surprisal for the continuations \textit{and} (\textit{or} is grammatical in all conditions, so it is omitted from this study). Our expectation is that `and' is less surprising in the \textit{Vpl\_Nsg} condition than in the \textit{Vsg\_Nsg} condition, where a CoordNP is not licensed by the grammar in either French or English (as in \textit{*What is the pig and the cat eating?}). We also expect lower surprisal for \textit{and} in the \textit{Vpl\_Nsg} condition, where it is obligatory for a grammatical continuation, than in the \textit{Vpl\_Npl} condition, where it is optional.

\begin{table}[t]
\centering
\small
\begin{tabular}{ll}
\toprule
{} Condition & Sentence preamble  \\
\midrule
Vpl\_Npl & What are the doors and \\
Vpl\_Nsg&  What are the door and \\
Vsg\_Nsg & What is the door and \\
\bottomrule
\end{tabular}
\caption{Conditions in Inverted Coordination experiment.}
\label{cond3}
\end{table}

For French experimental items, the question is embedded into a sentential-complement taking verb, following Example \ref{frenchinv}, due to the fact that unembedded subject-verb inverted questions sound very formal and might be relatively rare in the training data.

\exg. Je me demande où vont le maire et\\
I myself ask where go.3PL the.MSG mayor.MSG and\\  \label{frenchinv}

\begin{figure}[t]
    \centering
   \subfloat[English]{
    \includegraphics[width=0.9\linewidth]{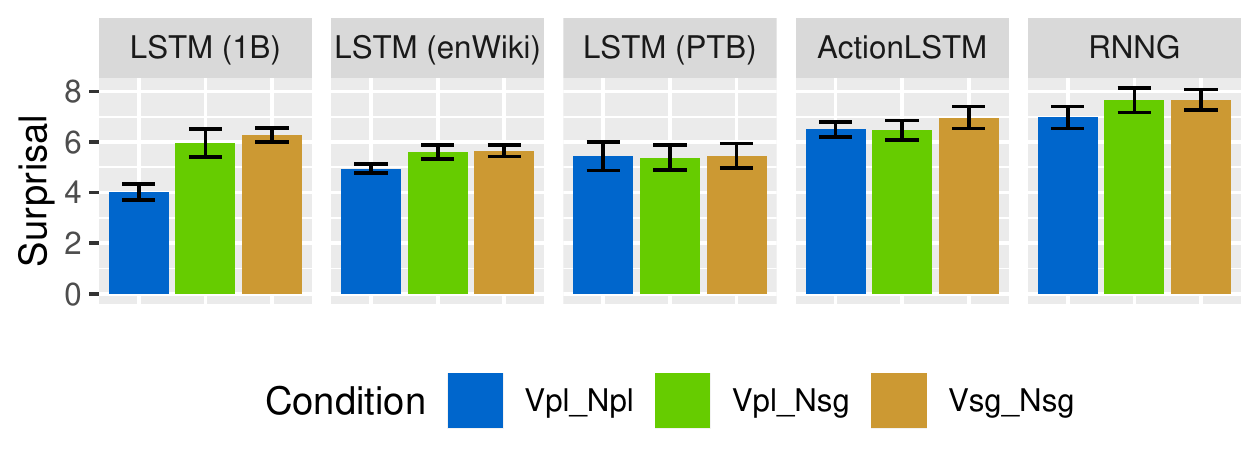}}
 
    \subfloat[French]{
    \includegraphics[width=0.9\linewidth]{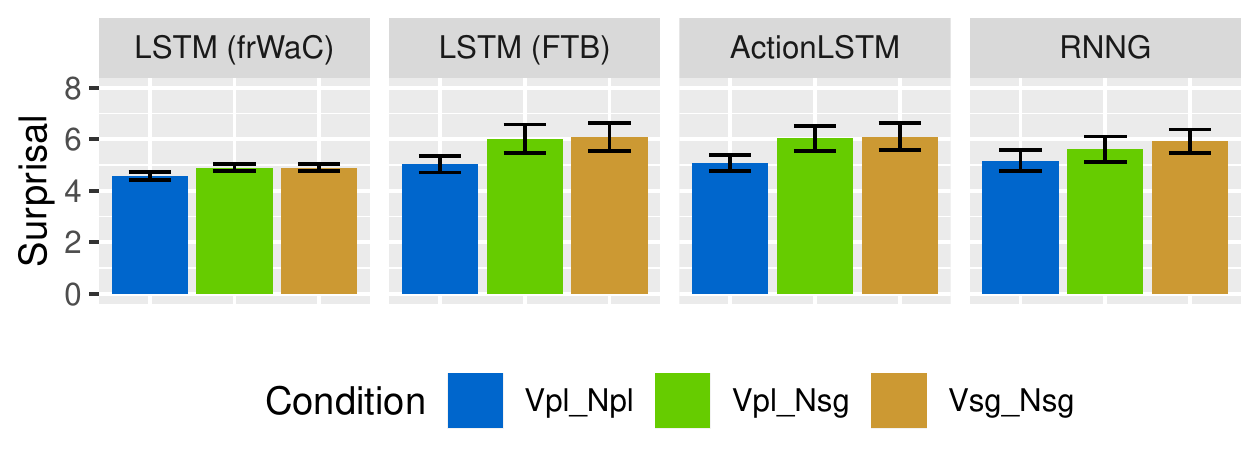}
    }
    \caption{Comparison of models' surprisals of 
    \textit{and}-coordination in Inverted Coordination experiment.}
    \label{inversion}
\end{figure}

The results for both languages are shown in Figure \ref{inversion}, with the surprisal at the coordinator on the y-axis and the various conditions on the x-axis. No model in either language shows a signficant difference in surprisal between the \textit{Vpl\_Nsg} and \textit{Vpl\_Npl} conditions or between the \textit{Vpl\_Nsg} and \textit{Vsg\_Nsg} conditions. The LSTM (1B) shows significant difference between the \textit{Vpl\_Nsg} and \textit{Vpl\_Npl} conditions, but in the opposite direction than expected, with the coordinator less surprising in the latter condition. These results indicate that the models are unable to use the fine-grained context sensitivity to drive expectations for CoordNPs, at least in the inversion setting.

\section{Discussion}

The experiments presented here extend and refine a line of research investigating what linguistic knowledge is acquired by neural language models. Previous studies have demonstrated that sequential models trained on a simple regime of optimizing the next word can learn long-distance syntactic dependencies in impressive detail. Our results provide complimentary insights, demonstrating that a range of model architectures trained on a variety of datasets can learn fine-grained information about the interaction of CoordNPs and local syntactic context, but their behavior remains unhumanlike in many key ways. Furthermore, to our best knowledge, this work presents the first psycholinguistic analysis of neural language models trained on French, a high-resource language that has so far been under-investigated in this line of research.

In the \textbf{simple coordination} experiment, we demonstrated that models were able to capture some of the agreement behaviors of humans, although their performance deviated in crucial aspects. Whereas human behavior is best modeled as a `percolation' process, the neural models appear to be using a linear combination of NP constituent number to drive CoordNP/verb number agreement, with the second noun weighted more heavily than the first. In these experiments, supervision afforded by the RNNG and ActionLSTM models did not translate into more robust or humanlike learning outcomes.
The \textbf{complex coordination} experiments provided evidence that the neural models tested were not using a simple `bag of features' strategy, but were sensitive to syntactic context. All models tested were able to interpret material that had similar surface form in ways that corresponded to two different tree-structural descriptions, based on local context. The \textbf{inverted coordination} experiment provided a contrasting example, in which models were unable to modulate expectations based on subtleties in the syntactic environment.

Across all our experiments, the French models performed consistently better on subject/verb number agreement than on subject/predicate gender agreement. Although there are likely more examples of subject/verb number agreement in the French training data, gender agreement is syntactically mandated and widespread in French. It remains an open question why all but one of the models tested were unable to leverage the numerous examples of gender agreement seen in various contexts during training to drive correct subject/predicate expectations.

\section*{Acknowledgments}

This project is supported by a grant of Labex EFL ANR-10-LABX-0083 (and Idex ANR-18-IDEX-0001)  for AA and MIT–IBM AI Laboratory and the MIT–SenseTimeAlliance on Artificial Intelligence for RPL. We would like to thank the anonymous reviewers for their comments and  Anne Abeillé for her advice and feedback.

\bibliography{emnlp-ijcnlp-2019}
\bibliographystyle{acl_natbib}

\appendix

\section{The Effect of Annotation Schemes}\label{appx:annotation}

\begin{figure}[t]
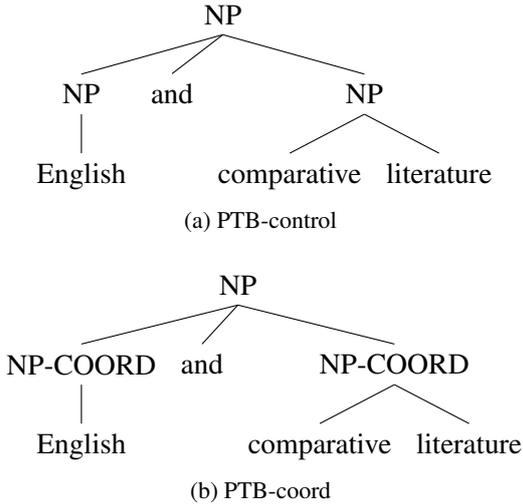

    \centering
    \subfloat[PTB-control]{
    \Tree [.NP [.NP [.English ] ] [.and ] [.NP [.comparative ] [.literature ] ] ]
    }
    
    \subfloat[PTB-coord]{
    \Tree [.NP [.NP-COORD [.English ] ] [.and ] [.NP-COORD [.comparative ] [.literature ] ] ]
    }
    \caption{Comparison of annotation schemes of coordination structure.}
    \label{tree-annotation}
\end{figure}

This section further investigates the effects of CoordNP annotation schemes on the behaviors of structurally-supervised models. We test whether an explicit COORD phrasal tag improves model performance. We trained two additional RNNG models on 38,546 sentences from the Penn Treebank  annotated with two different schemes: The first, \textbf{RNNG (PTB-control)} was trained with the original Penn Treebank annotation. The second, {\bf RNNG (PTB-coord)}, was trained on the same sentences, but with an extended coordination annotation scheme, meant to employ the scheme employed in the FTB, adapted from \citet{ficler2016coordination}. We stripped empty categories from their scheme and only kept the NP-COORD label for constituents inside a coordination structure. Figure \ref{tree-annotation} illustrates the detailed annotation differences between two datasets. We tested both models on all the experiments presented in Sections \ref{sec:control}-\ref{sec:inversion} above.

Turning to the results of these six experiments: We see little difference between the two models in the \textit{Non-coordination agreement} experiment. For the \textit{Complex coordination control} and \textit{Complex coordination critical} experiments, both models are largely the same as well. However, in the \textit{Simple and-coordination} and \textit{Simple or-coordination} experiments the values for all conditions are shifted upwards for the RNNG PTB-coord model, indicating higher over-all preference for the plural continuation. Furthermore, the range of values is reduced in the RNNG PTB-coord model, compared to the RNNG PTB-control model. These results indicate that adding an explicit COORD phrasal label does not drastically change model performance: Both models still appear to be using a linear combination of number features to drive plural vs. singular expectation. However, the explicit representation has made the interior of the coordination phrase more opaque to the model (each feature matters less) and has slightly shifted model preference towards plural continuations. In this sense, the PTB-coord model may have learned a generalization about CoordNPs, but this generalization remains unlike the ones learned by humans.

\begin{figure*}[!t]
 \subfloat[Non-coordination agreement]{\includegraphics[width=0.33\linewidth]{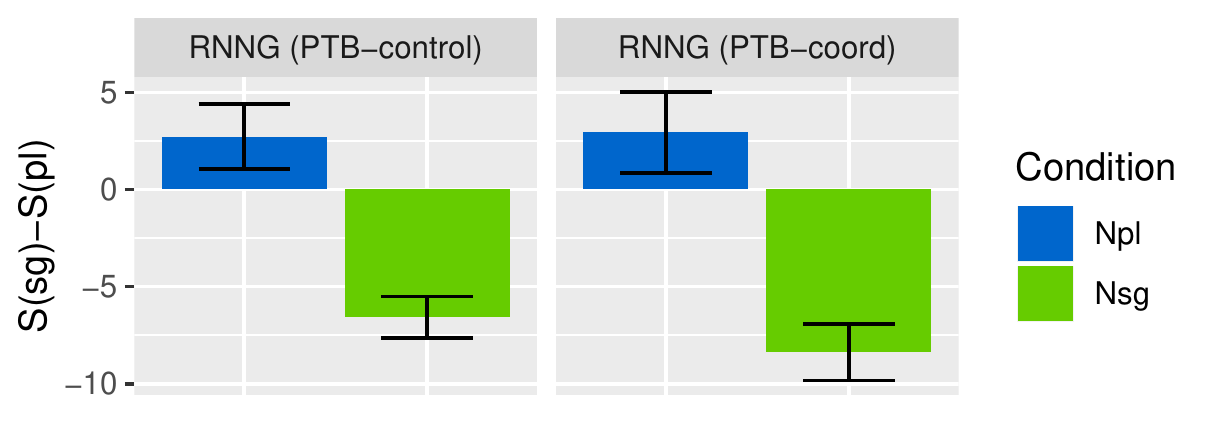}}
    \subfloat[Simple and-coordination]{\includegraphics[width=0.33\linewidth]{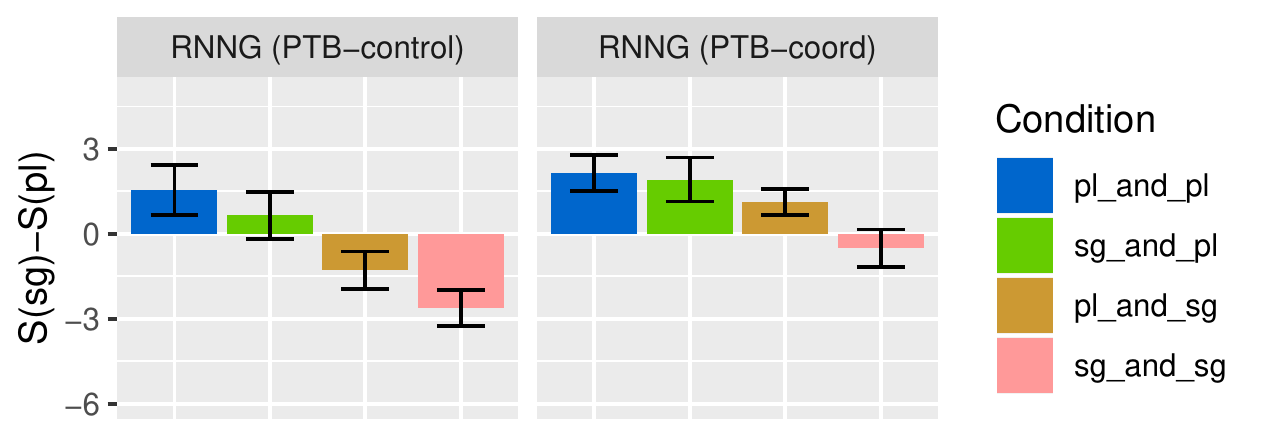}}
    \subfloat[Simple or-coordination]{\includegraphics[width=0.33\linewidth]{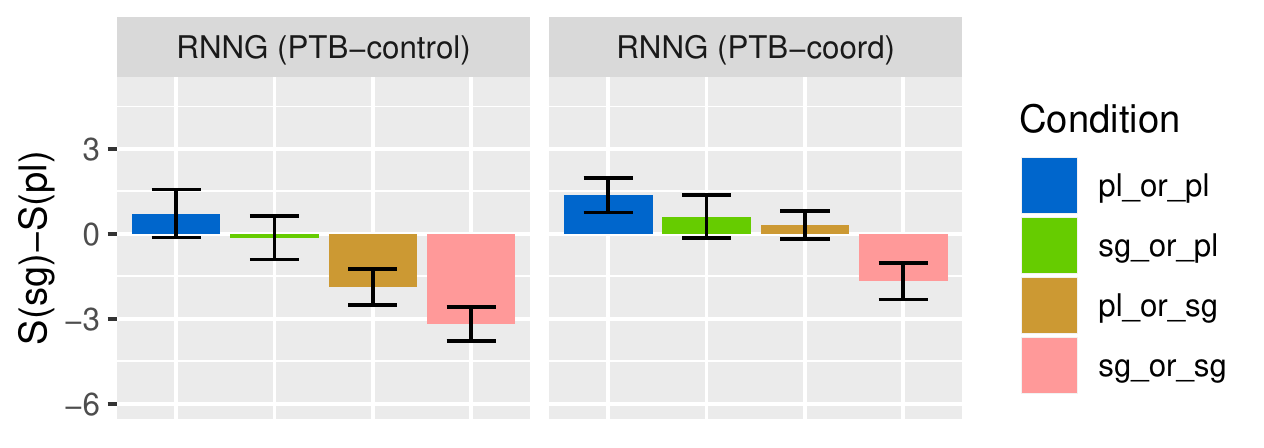}}  
    
   \subfloat[Complex coordination control]{\includegraphics[width=0.33\linewidth]{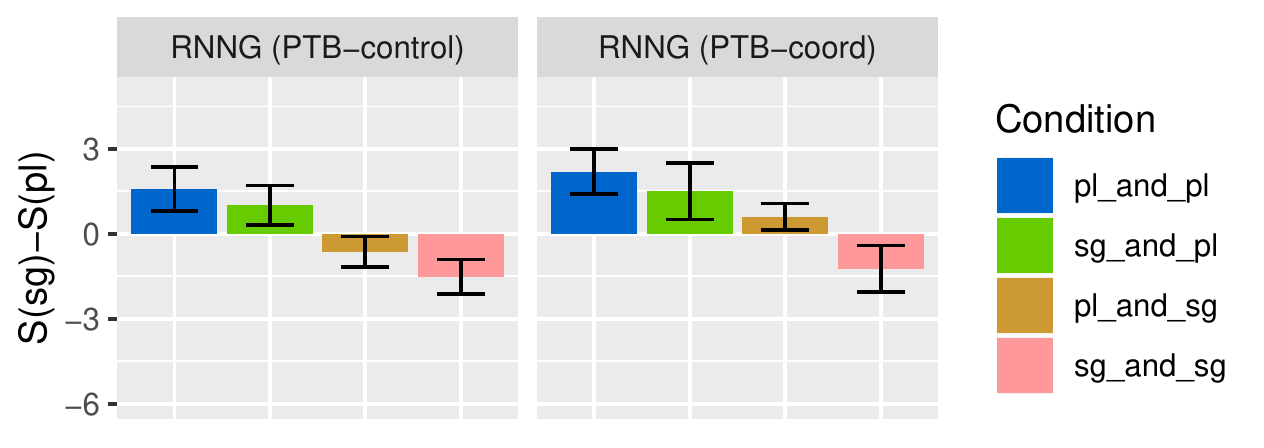}}
   \subfloat[Complex coordination critical]{\includegraphics[width=0.33\linewidth]{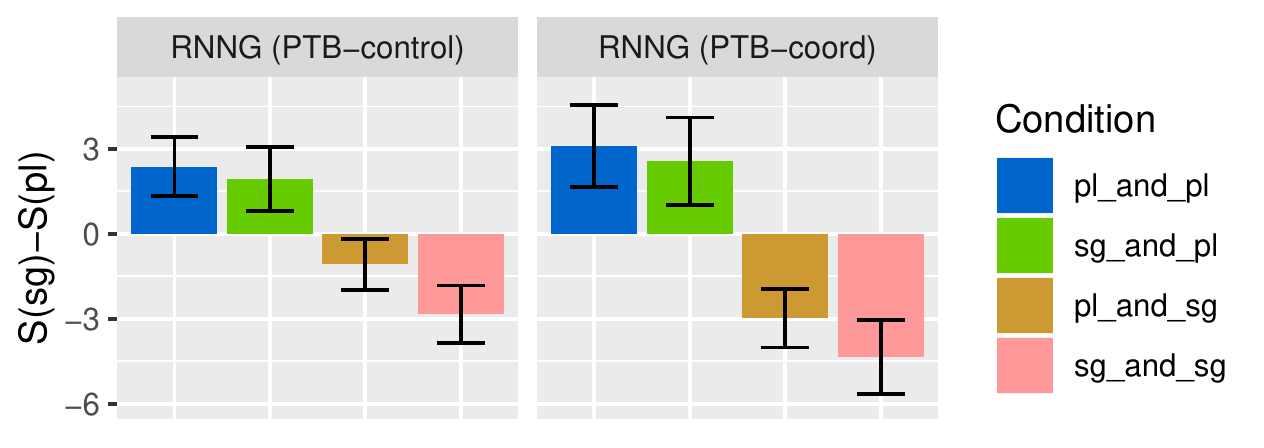}}
    \subfloat[Inverted coordination]{\includegraphics[width=0.33\linewidth]{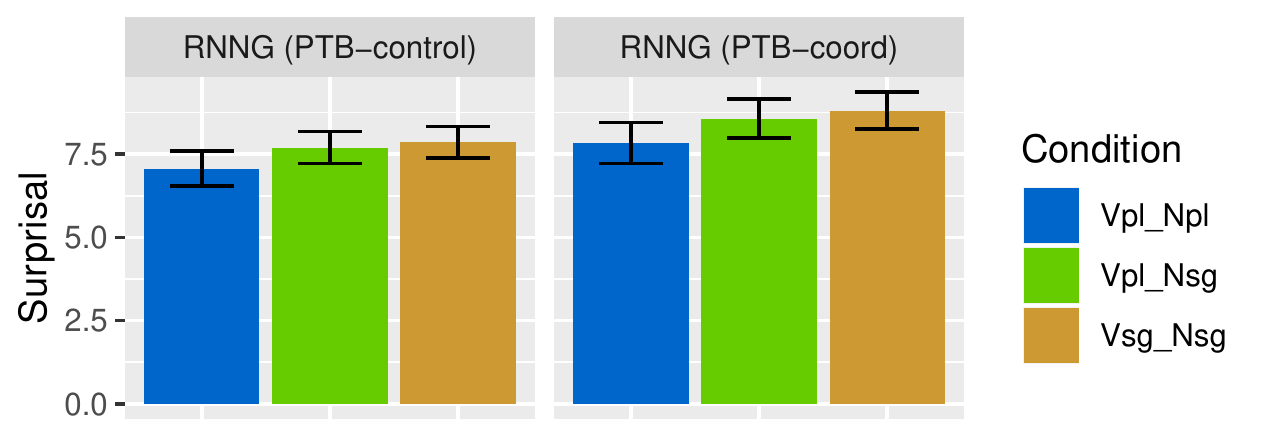}}
    \caption{Comparison between RNNGs trained on PTB data with original annotation vs. fine-grained annotation of coordination structure.}
    
    \label{compare-control-coord}
\end{figure*}

\begin{table}[!t]
\centering
\small
\begin{tabular}{l|rrr|rrr}
\toprule
& \multicolumn{3}{|c|}{PTB} & \multicolumn{3}{|c}{FTB} \\
\midrule
Condition & sg & pl& total & sg & pl & total  \\
\midrule
pl\_and\_pl & 0 &67 & 67&1&116&116\\
sg\_and\_pl &0 &72&72 &0& 50&50\\
pl\_and\_sg & 0 & 11 & 11&0 &30&30\\
sg\_and\_sg & 7 &213 & 220&5&288&293\\
pl\_or\_pl & 0 &2&2 &0&14 &14\\
sg\_or\_pl & 0& 0& 0&0&0&0\\
pl\_or\_sg & 0&1 &1&0&1 &1\\
sg\_or\_sg & 5&1&6&5&8&13\\

\bottomrule
 \end{tabular}
 \caption{Frequency of number agreement patterns in PTB and FTB.}
 \label{tab:conditionexp1-stats-number}
 \end{table}
 
 \begin{table}[!t]
\centering
\small
\begin{tabular}{lrrr}
\toprule
{} Condition & m & f& total  \\
\midrule
m\_and\_m & 38 &0 & 38\\
m\_and\_f &10 &1&11 \\
f\_and\_m & 9 & 0 & 9\\
f\_and\_f & 0 &16 &16 \\
m\_or\_m & 1 &0 & 1\\
m\_or\_f &0 &0&0 \\
f\_or\_m & 1 & 0 & 1\\
f\_or\_f & 0 &1&1 \\

\bottomrule
 \end{tabular}
 \caption{Frequency of gender agreement patterns in FTB.}
 \label{tab:conditionexp1-stats-gender}
 \end{table}

\section{PTB/FTB Agreement Patterns }

We present statistics of subject/predicate agreement patterns in the Penn Treebank (PTB) and French Treebank (FTB) in Table \ref{tab:conditionexp1-stats-number} and \ref{tab:conditionexp1-stats-gender}.

\end{document}